\pdfoutput=1

\documentclass[11pt]{article}

\usepackage[final]{acl}

\usepackage{times}
\usepackage{latexsym}

\usepackage[T1]{fontenc}

\usepackage[utf8]{inputenc}

\usepackage{microtype}

\usepackage{float}
\usepackage{inconsolata}
\usepackage{makecell}
\usepackage{mathrsfs}
\usepackage{tabularx}
\usepackage{threeparttable}
\usepackage{amsthm}
\usepackage{amsmath,amsfonts,bm}
\usepackage{multicol}
\usepackage{multirow}
\usepackage{booktabs}
\usepackage{graphicx}
\usepackage{mdframed}
\usepackage{bold-extra}
\usepackage{xspace}
\usepackage{xcolor}
\definecolor{darkgreen}{rgb}{0,0.6,0}
\usepackage{colortbl}
\usepackage{tcolorbox}
\usepackage{listings}
\usepackage{multicol}
\usepackage{lipsum}
\usepackage{enumitem}
\newcommand{\grayline}{\rowcolor[gray]{.90}}

\newcommand{\model}{Synthesizer\xspace}

\title{CoT-based Synthesizer: Enhancing \\ LLM Performance through Answer Synthesis}

\author{
Bohan Zhang$^{1,3}$\thanks{Equal Contributions.}\thanks{Work was done when interned at Zhipu AI.},
Xiaokang Zhang$^{1,3}$\footnotemark[1],
Jing Zhang$^{1,3}$\thanks{Corresponding Author.},
Jifan Yu$^2$,
Sijia Luo$^1$,
Jie Tang$^2$
\\[3pt]
$^1$School of Information, Renmin University of China,
$^2$Tsinghua University,
\\
$^3$Key Laboratory of Data Engineering and Knowledge Engineering, Beijing, China
\\
\{zbhmint, zhang2718, zhang-jing\}@ruc.edu.cn\\
\vspace{-4ex}
}

\begin{document}

\maketitle

\begin{abstract}
Current inference scaling methods, such as Self-consistency and Best-of-N, have proven effective in improving the accuracy of LLMs on complex reasoning tasks. 
However, these methods rely heavily on the quality of candidate responses and are unable to produce correct answers when all candidates are incorrect.
In this paper, we propose a novel inference scaling strategy, CoT-based \model, which leverages CoT reasoning to synthesize superior answers by analyzing complementary information from multiple candidate responses, even when all candidate responses are flawed. 
To enable a lightweight and cost-effective implementation, we introduce an automated data generation pipeline that creates diverse training data. This allows smaller LLMs trained on this data to improve the inference accuracy of larger models, including API-based LLMs.
Experimental results across four benchmark datasets with seven policy models demonstrate that our method significantly enhances performance, with gains of 11.8\% for Llama3-8B and 10.3\% for GPT-4o on the MATH500 dataset. The corresponding training data and code are publicly available on the repository\footnote{\url{https://github.com/RUCKBReasoning/CoT-based-Synthesizer}}.
\end{abstract}


\section{Introduction}
While large language models (LLMs)~\cite{NEURIPS2020_1457c0d6,Aakanksha2023palm,touvron2023llama,achiam2023gpt} have achieved revolutionary progress in the field of natural language processing (NLP), they often struggle to generate accurate answers in one attempt when dealing with complex reasoning tasks~\cite{rae2021scaling,he2024olympiadbench,tong2024dart}. To address this, researchers have focused on expanding the scale of inference to enhance model inference performance~\cite{nye2021show,wu2024empirical,brown2024large}.

\begin{figure}[t]
    \centering
    \includegraphics[width=\linewidth]{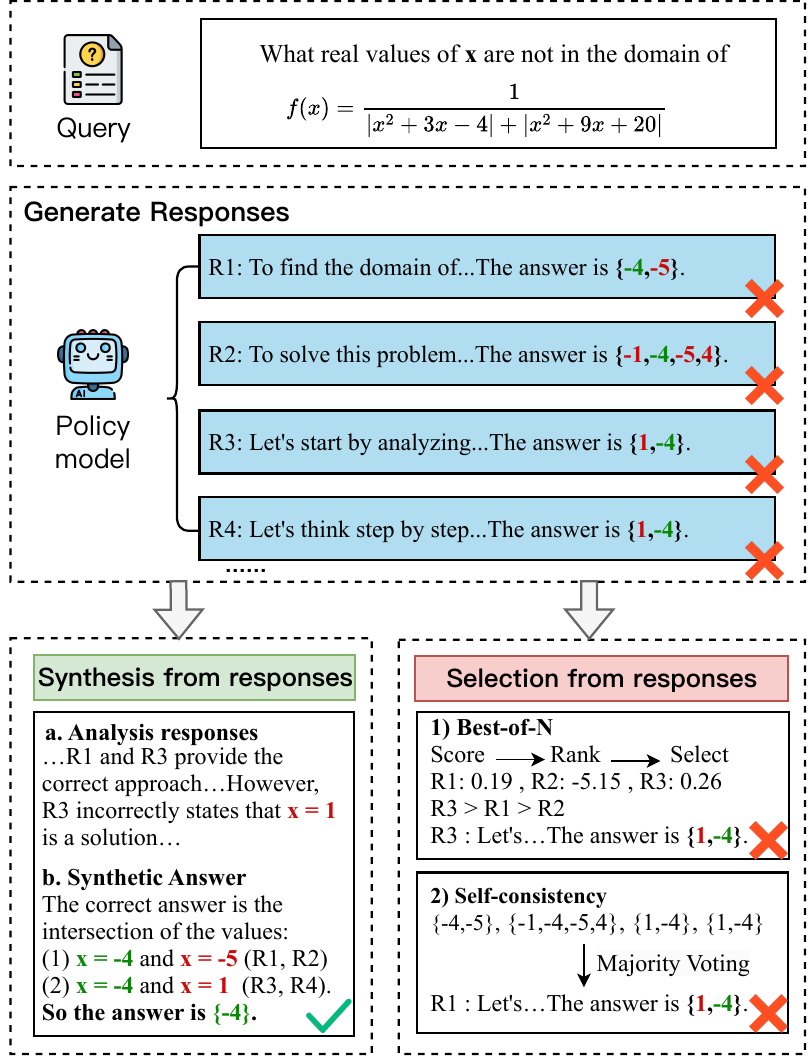}
    \caption{An example of our method in mathematical reasoning. Even when the policy model generates all incorrect responses, our method can still leverage their strengths to produce the correct answer.}
    \label{fig:showcase}
\end{figure}

A common inference scaling strategy is Best-of-N (BoN), which involves training a verifier to score multiple candidate responses generated by the model and selecting the highest-scoring answer as the final result~\cite{stiennon2020learning,cobbe2021training,lightman2024lets}.
However, the scoring process for each candidate response in Best-of-N is independent, which fails to leverage the potential relationships between the candidates. This may lead to insufficient information integration and a susceptibility to reward hacking~\cite{skalse2022defining,singhal2023long}.
Since candidates are evaluated in isolation, the lack of cross-sample comparison makes it hard to detect outputs that exploit flaws in the reward function~\cite{christiano2017deep,ouyang2022training}. 
Another widely used inference scaling strategy is Self-consistency (SC), which takes a distinct approach by selecting the most frequently generated answer from multiple inferences as the final result~\cite{wang2022self}. Unlike BoN, SC does not require an additional scoring step by a verifier. Instead, it uses a majority-voting process based on exact matches of candidate responses. However, this reliance on exact matching limits SC’s applicability in open-ended tasks. To address this limitation, the Universal Self-Consistency (USC)~\cite{chen2023universal} leverages the generative capabilities of LLMs for voting, replacing exact matching in SC, thus expanding its range of applications. However, when the correct answer appears at a low frequency among the candidates, consistency methods may fail to select the correct answer. In summary, both BoN and consistency methods operate on the assumption that the correct answer exists within the candidate set. When all candidate responses are incorrect, these methods fail to generate the correct answer.
Given that BoN and USC both rely on post-processing models to evaluate candidate responses, a natural question emerges: \textit{Can we go beyond selecting from existing candidates and instead synthesize a better answer by combining the strengths of multiple candidate responses?}

Based on this insight, we propose a novel inference scaling strategy: \textbf{CoT-based \model}, which leverages Chain-of-Thought (CoT)~\cite{wei2022chain} reasoning to systematically analyze candidate responses and synthesize new answers during the inference process. By doing so, the model can identify the strengths of each candidate and synthesize them into a more accurate and complete final answer. As illustrated in Figure~\ref{fig:showcase}, Best-of-N and Self-consistency fail to derive the correct answer from flawed candidates. In contrast, by encouraging the model to deeply analyze and integrate information across candidate responses, our method enables the synthesis of the correct answer. 
To enable training a smaller yet effective LLM for this purpose, we further propose an automated data generation pipeline designed to create training data for candidate response synthesis. By sampling multiple inference outputs from LLMs and filtering to ensure quality, we collect diverse data with complementary information for training. 

We conduct experiments on 4 benchmarks and evaluate our method using 7 models with varying parameter scales for candidate response generation. The results demonstrate that small models trained using the data generation pipeline can not only effectively enhance the performance of candidate responses generated by various large models, even API models (e.g., with gains of $11.8\%$ for Llama3-8B and $10.3\%$ for GPT-4o on MATH500), but also outperform traditional methods such as Self-consistency and Best-of-N. 

Our contributions can be summarized as follows: 
1) We propose a novel inference scaling strategy: CoT-based \model, which synthesizes better results by analyzing candidate responses.
2) We design an automated data generation pipeline to enable training smaller efficient models for superior answer synthesis.
3) We validate the effectiveness of our method across multiple NLP tasks, achieving significant improvements over existing methods in complex reasoning scenarios.



\section{Related Work}
\paragraph{Best-of-N.} The Best-of-N approach, introduced as a common method for improving the quality of generated outputs, involves sampling multiple solutions and selecting the best one based on a scoring mechanism~\cite{stiennon2020learning}. Previous works have trained a reward model, often referred to as a ``verifier'', to discriminate between correct and incorrect solutions, either through binary classification~\cite{cobbe2021training,uesato2022solving,yu2023outcome} or by ranking preferences between solutions~\cite{stiennon2020learning,nakano2021webgpt}. Recent works integrate the generative capabilities of LLMs with the discriminative power of reward models, enabling both scoring and explaining the scoring process~\cite{zhang2024generative,yu2024self}. This paper focuses exclusively on outcome-based methods, leaving process-oriented approaches beyond the scope of discussion.

\paragraph{Consistency Methods.} Self-consistency is another common inference scaling method, which operates under the assumption that the most consistent answer, determined through a voting process, is more likely to be correct. Methods based on consistency have demonstrated significant performance improvements across various domains like mathematics~\cite{wang2022self}, code generation~\cite{shi2022natural,li2022competition} and open-ended question answering~\cite{chen2023universal}. Our method shares similarities with~\cite{chen2023universal} in utilizing LLMs for inference scaling, while our method is distinguished by its emphasis on the generation of novel answers. Notably, even when all candidate responses are incorrect, our method retains the capability to yield accurate outcomes.

\paragraph{Inference Scaling Synthesis Methods.}
Recent studies~\cite{farinhas2023empirical,vernikos2023small} have explored answer synthesis by combining candidate responses. Additionally,~\cite{vernikos2024don} proposes methods that employ quality estimation metrics to effectively combine outputs from LLMs. These approaches have successfully validated the efficacy of synthesizing high-quality responses in translation tasks. Similarly, ensemble methods have been explored for combining candidate responses from different LLMs. One such method trains a ranker to conduct pairwise comparisons and selects suitable candidates to fuse into a superior answer~\cite{jiang2023llm}. However, our work focuses on synthesizing candidate responses generated exclusively by a single LLM without using diverse models. 


\section{Problem Formulation}
This study concentrates on post-processing multiple candidate responses generated by the policy model to produce a new synthesized answer.
In this paper, the policy model refers to the model that generates direct responses to user queries.
Formally, given a user query \(x\), the policy model performs \(N\) rounds of random sampling-based generation from a probability distribution, producing a set of responses denoted as: 
\(R = \{r_1, r_2, \dots, r_N\}\),
where $r_i$ represents the $i$-th candidate answer. These responses, together with the query \(x\), are then combined into \(\{x, R\}\). Subsequently, an analysis and synthesis model, referred to as the \model, is employed to regenerate a new answer $y$.
This process can be formalized as:

\[
(x, R) \xrightarrow{f_{\text{\model}}} y
\]
Here, \(f_{\text{\model}}\) denotes the synthesis function applied to the query \(x\) and candidate set \(R\) to produce the final answer \(y\).




\section{Methodology}
\begin{figure*}[t]
  \centering
  \includegraphics[width=\linewidth]{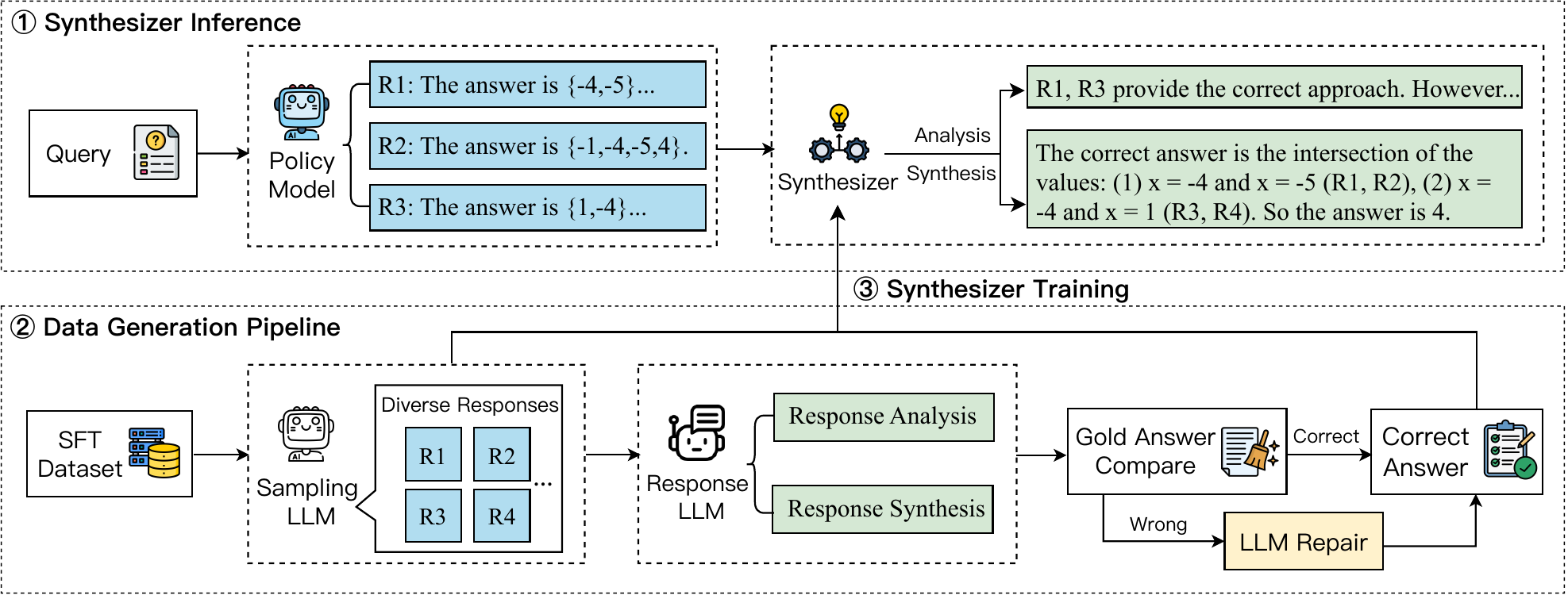}
  \caption{An overview of our method. (1) \textbf{Synthesizer Inference}: The policy model generates diverse candidate responses, which are analyzed and synthesized by the CoT-based synthesizer to produce a high-quality final response. (2) \textbf{Data Generation Pipeline:} The pipeline combines a diverse response generation process using a sampling LLM with query-response relationship analysis to construct high-quality synthetic data. (3) \textbf{Synthesizer Training:} The generated dataset is then used to train the model via SFT to enhance reasoning and synthesis capabilities.}
  \label{fig:overview fig}
\end{figure*}

As illustrated in Figure~\ref{fig:overview fig}, we present an overview of our methodology. Specifically, we propose an inference scaling strategy based on the CoT reasoning, enabling the model to analyze and synthesize information across multiple candidate responses. To enable training a smaller yet effective LLM, we devise a two-stage data generation pipeline to construct a high-quality training dataset, which combines automated candidate generation with an extra repair step to improve data reliability. Finally, leveraging the generated dataset, we train \model-8B that integrates CoT reasoning to synthesize the final answer. Furthermore, our approach incorporates various LLMs, including the sampling LLM, response LLM, policy model, and base model, each playing a distinct role in inference and training. Formal definitions of these components are provided in Appendix~\ref{app:glossary}.

\subsection{\model Inference}

\paragraph{Diverse Response Generation.}
\label{sub_sec:response_generation}
We use the query as the input for the policy model to generate a set of candidate answers. To ensure diversity and quality among candidates, we employ the following decoding strategies during generation: (1) We set a relatively high sampling temperature ($t=0.9$) to increase randomness and promote diversity. (2) We employ Top-P ($p=0.9$) sampling decoding~\cite{Holtzman2020The}, which truncates the candidate pool to include only the most probable tokens up to a cumulative probability of $p$. This process inherently avoids generating rare but semantically meaningless low-probability candidates, ensuring that the generated responses remain coherent and relevant. By combining these strategies, we generate a diverse yet coherent ensemble of $N$ responses, designated as $(x, R)$.

\paragraph{Response Analysis and Synthesis.}
\label{sub_sec:response_synthesis}
We input the generated $(x, R)$ into the CoT-based \model, which aims to analyze candidate responses, extract accurate information, and ultimately produce a high-quality response. The process involves the following steps: (1) \textbf{Response Analyis}: \model first conducts a thorough analysis of the relationship between the user query and each candidate response. It takes into account the frequency, relevance, and accuracy of similar responses to user queries. While high-frequency answers may indicate higher credibility, the \model does not overly rely on frequency as an evaluation criterion. Instead, it prioritizes logical coherence and factual accuracy, making it possible to identify even partially correct or misleading answers and extract valuable information from them. (2) \textbf{Response Synthesis}: When a correct answer exists within the candidate set, the \model further enhances and enriches the final response by incorporating effective reasoning steps from other candidate answers. In cases where there is a lack of clearly correct options or all candidate answers have deficiencies, the \model leverages its reasoning capabilities to integrate reasonable elements from multiple candidates, constructing a more coherent and accurate answer.
These processes are achieved by designing specific prompts for the \model, enabling structured reasoning and accurate response generation. Detailed information can be seen in Box~\ref{box:synthesizer_inference}.

\subsection{Data Generation Pipeline}
\label{sub_sec:Training_data_preparation}
To enable the training of CoT-based \model, we design a two-stage data generation pipeline. Our pipeline is built on top of existing benchmarks, where we denote the query in the benchmark as $x$ and its corresponding gold answer as $y_g$. 

\paragraph{Synthetic Answer Generation.}
For a given training query $x$, we follow the diverse response generation process described in Section~\ref{sub_sec:response_generation}, using a sampling LLM (e.g., Llama3-8B-Instruct) to obtain the pair $(x, R)$. Then we input the pair into a high-performance response LLM (e.g., Llama3.1-70B-Instruct), which employs a CoT reasoning process to generate responses. Given the complexity of $x$, the response LLM may not synthesize an accurate answer on the first attempt. To address this problem, we perform multiple samplings, significantly increasing the likelihood of obtaining a correct response~\cite{tong2024dart}. Specifically, we sample the response LLM $N$(e.g., $N=50$) times to conduct response analysis and synthesis as described in Section~\ref{sub_sec:response_synthesis}, generating a diverse set of synthetic answers. To ensure the quality of these synthetic answers, we filter responses using the gold answer $y_g$, retaining correct responses $y$, which contain CoT analysis and synthesis of $R$.

\paragraph{LLM Repair.}
In cases where all candidate responses $R$ are incorrect, it is challenging for the response model to generate a correctly synthesized answer during the initial sampling stage. Specifically, we explicitly inform the response model that all candidate responses $R$ are incorrect and prompt it to reflect on the errors in these responses. By analyzing the incorrect answers, the model identifies valid reasoning steps and synthesizes a refined response that is more likely to be correct. Then we filter the newly generated responses to obtain the correct responses $y$.

\subsection{Synthesizer Training}
Through the above stages, we construct the synthesis training dataset $M = \{(x^{(i)}, R^{(i)}, y^{(i)})\}_{i=1}^N$, where
$x$ represents the user query, 
$R$ is the set of candidate responses, and $y$ is the correct synthesis answer generated by the Response LLM. Based on this dataset, we train the model CoT-based \model-8B. The model takes the user query $x$ and the candidate response set $R$ as input and generates the correct answer $y$ as output. The generation probability is defined as:
\begin{equation}
    p_\phi(y \mid x, R) = \prod_{i=1}^T p_\phi(y_i \mid x, R, y_{<i}),
\end{equation}
where $T$ is the sequence length, $y_i$ is the $i$-th target token, and $y_{<i}$ represents the generated context (i.e., all previous tokens). 

Through training, the model learns to leverage both the user query $x$ and the candidate response set $R$ to generate accurate and contextually appropriate answers $y$. This process enhances the model's ability to synthesize coherent outputs while effectively utilizing the provided candidates.

\section{Experiment}

\begin{table*}[t]
    \centering
    \caption{The main result of our approach and other baselines over GSM8k, MATH500, WikiTQ and FeTaQA. The top two performances are highlighted in \textbf{bold} and \underline{underlined}.}
    \label{tab:Main results}
    \resizebox{\textwidth}{!}{%
    \begin{tabular}{lcccccccc}
    \toprule
    \multirow{2}{*}[-3pt]{Method} &  
    \multicolumn{7}{c}{Policy Models} \\ \cmidrule{2-9}
    & GLM-4-Plus & GPT-4o & Llama3.1-70B & Llama3.1-8B & Llama3-8B & Qwen2-7B & Qwen2.5-14B & Average \\
    \midrule
    \grayline \multicolumn{9}{c}{\textbf{GSM8k}} \\
    \midrule
    CoT-prompting~\cite{wei2022chain} & 88.6 & 91.4 & 92.7 & 81.9 & 73.4 & 82.0 & 91.2 & 85.9 \\
    SC~\cite{wang2022self} & 90.1 & 92.4 & \underline{93.9} & 85.1 & 80.9 & 84.3 & \underline{92.3} & 88.4 \\
    USC (Llama3.1-70B)~\cite{chen2023universal} & 90.1 & 92.3 & 93.5 & 85.4 & 82.0 & 84.9 & \underline{92.3} & 88.6 \\
    LMCOR (Llama3-8B)~\cite{vernikos2023small} & 88.9 & 90.4 & 90.1 & 83.1 & 79.4 & 84.8 & 89.5 & 86.6 \\
    ArmoRM (Llama3-8B)~\cite{wang2024interpretable} & 90.3 & 91.6 & 93.3 & 85.5 & \underline{82.4} & 86.1 & 92.1 & 88.8 \\
    Scalar RM (Llama3-8B)~\cite{cobbe2021training} & 89.1 & 91.9 & 93.3 & 85.6 & 81.6 & 85.8 & 91.4 & 88.4 \\
    \midrule
    ours (Llama3.1-70B) & \underline{91.2} & \underline{92.6} & 93.6 & \textbf{86.9} & \textbf{83.5} & \textbf{88.3} & \underline{92.3} & \textbf{89.8} \\
    ours (\model-8B) & \textbf{91.4} & \textbf{93.0} & \textbf{94.0} & \underline{86.1} & 81.3 & \underline{86.4} & \textbf{92.7} & \underline{89.3} \\
    \midrule
    \grayline \multicolumn{9}{c}{\textbf{MATH500}} \\
    \midrule
    CoT-prompting~\cite{wei2022chain} & 54.8 & 62.5 & 66.6 & 46.5 & 24.2 & 57.3 & 74.4 & 55.2 \\
    SC~\cite{wang2022self} & 63.0 & 68.7 & 68.8 & \textbf{55.4} & 32.4 & 61.0 & 76.6 & 60.8 \\
    USC (Llama3.1-70B)~\cite{chen2023universal} & 62.6 & 67.3 & 68.4 & 52.8 & 35.4 & 62.2 & \underline{78.2} & 61.0 \\
    LMCOR (Llama3-8B)~\cite{vernikos2023small} & 52.4 & 61.2 & 57.6 & 44.8 & 33.6 & 51.6 & 64.0 & 52.2 \\
    ArmoRM (Llama3-8B)~\cite{wang2024interpretable} & 60.6 & 67.5 & \underline{69.4} & 52.6 & 32.8 & 60.2 & 77.2 & 60.0 \\
    Scalar RM (Llama3-8B)~\cite{cobbe2021training} & 61.4 & 65.9 & 66.8 & 52.8 & 34.2 & 59.4 & 77.6 & 59.7 \\
    \midrule
    ours (Llama3.1-70B) & \underline{64.2} & \textbf{75.5} & \textbf{69.6} & 52.8 & \textbf{38.8} & \textbf{63.6} & \textbf{79.0} & \textbf{63.4} \\
    ours (\model-8B) & \textbf{64.4} & \underline{72.8} & \textbf{69.6} & \underline{54.6} & \underline{36.0} & \underline{62.4} & \underline{78.2} & \underline{62.6} \\
    \midrule
    \grayline \multicolumn{9}{c}{\textbf{WikiTQ}} \\
    \midrule
    CoT-prompting~\cite{wei2022chain} & 90.1 & 89.9 & 86.7 & 72.4 & 71.7 & 63.8 & 77.9 & 78.9 \\
    USC (Llama3.1-70B)~\cite{chen2023universal} & 91.6 & 91.8 & \underline{88.3} & 79.6 & 76.3 & 69.2 & 81.5 & 82.6 \\
    LMCOR (Llama3-8B)~\cite{vernikos2023small} & 88.8 & 90.4 & 87.8 & 77.3 & 75.4 & 69.2 & 81.4 & 81.5 \\
    ArmoRM (Llama3-8B)~\cite{wang2024interpretable} & 91.0 & 91.8 & 87.5 & 77.9 & 73.8 & 69.4 & 81.2 & 81.8 \\
    Scalar RM (Llama3-8B)~\cite{cobbe2021training} & 91.8 & 90.5 & 87.5 & 77.6 & 74.9 & 69.8 & 80.1 & 81.7 \\
    \midrule
    ours (Llama3.1-70B) & \underline{91.9} & \textbf{92.3} & \underline{88.3} & \textbf{83.4} & \textbf{82.2} & \textbf{78.0} & \textbf{84.2} & \textbf{85.8} \\
    
    ours (\model-8B) & \textbf{92.1} & \underline{91.9} & \textbf{88.9} & \underline{79.9} & \underline{77.7} & \underline{72.2} & \underline{82.4} & \underline{83.6} \\
    \midrule
    \grayline \multicolumn{9}{c}{\textbf{FeTaQA}} \\
    \midrule
    CoT-prompting~\cite{wei2022chain} & 86.4 & 86.3 & 85.6 & 82.6 & 82.2 & 73.5 & 82.7 & 82.8 \\
    USC (Llama3.1-70B)~\cite{chen2023universal} & \underline{87.1} & \underline{87.0} & 86.1 & 84.3 & 83.9 & 77.5 & \underline{84.1} & 84.3 \\
    LMCOR (Llama3-8B)~\cite{vernikos2023small} & 86.0 & 84.7 & 83.0 & 84.7 & 83.8 & 79.9 & 83.2 & 83.6 \\
    ArmoRM (Llama3-8B)~\cite{wang2024interpretable} & \textbf{87.5} & 86.1 & 86.0 & 83.2 & 82.5 & 76.1 & 82.9 & 83.5 \\
    Scalar RM (Llama3-8B)~\cite{cobbe2021training} & 87.4 & 85.5 & 85.3 & 83.0 & 82.3 & 75.1 & 83.3 & 83.1 \\
    \midrule
    ours (Llama3.1-70B) & 87.0 & 86.8 & \underline{86.6} & \textbf{84.9} & \underline{85.1} & \textbf{82.3} & \underline{84.1} & \underline{85.3} \\
    ours (\model-8B) & \textbf{87.5} & \textbf{87.9} & \textbf{87.5} & \underline{84.7} &  \textbf{85.9} & \underline{82.1} & \textbf{86.6} & \textbf{86.0} \\

    \bottomrule
    \end{tabular}
    }

\end{table*}

\begin{table*}[ht]
\centering
\caption{The impact of the data generation pipeline and the process of CoT analysis of candidate responses. The improvements are
calculated between methods and CoT-prompting.}
\label{tab:ablation results}
\resizebox{\textwidth}{!}{%
\begin{tabular}{lllllllll}
\toprule
\multirow{2}{*} & & \multicolumn{7}{c}{Policy Models} \\ \cmidrule{2-9}
Method & GLM-4-Plus & GPT-4o & Llama3.1-70B & Llama3.1-8B & Llama3-8B & Qwen2-7B & Qwen2.5-14B & Average \\
\midrule
\grayline \multicolumn{9}{c}{\textbf{GSM8k}} \\
\midrule
CoT-prompting & 88.6 & 91.4 & 92.7 & 81.9 & 73.4 & 82.0 & 91.2 & 85.9 \\
\model-8B(ours) & 91.4\textsubscript{ +2.8} & 93.0\textsubscript{ +1.6} & 94.0\textsubscript{ +1.3} & 86.1\textsubscript{ +4.2} & 81.3\textsubscript{ +7.9} & 86.4\textsubscript{ +4.4} & 92.7\textsubscript{ +1.5} & 89.3\textsubscript{ +3.4} \\
\hspace{1em}w/o CoT training & 90.0\textsubscript{ +1.4} & 90.6\textsubscript{ -0.8} & 91.6\textsubscript{ -1.1} & 83.3\textsubscript{ +1.4} & 78.4\textsubscript{ +5.0} & 84.3\textsubscript{ +2.3} & 90.4\textsubscript{ -0.8} & 86.9\textsubscript{ +1.0} \\
\hspace{1em}w/o training & 87.0\textsubscript{ -1.6} & 87.3\textsubscript{ -4.1} & 89.4\textsubscript{ -3.3} & 80.1\textsubscript{ -1.8} & 75.4\textsubscript{ +2.0} & 81.8\textsubscript{ -0.2} & 87.5\textsubscript{ -3.7} & 84.1\textsubscript{ -1.8} \\
\midrule
\grayline \multicolumn{9}{c}{\textbf{MATH}} \\
\midrule
CoT-prompting & 54.8 & 62.5 & 66.6 & 46.5 & 24.2 & 57.3 & 74.4 & 55.2 \\
\model-8B(ours) & 64.4\textsubscript{ +9.6} & 72.8\textsubscript{ +10.3} & 69.6\textsubscript{ +3.0} & 54.6\textsubscript{ +8.1} & 36.0\textsubscript{ +11.8} & 62.4\textsubscript{ +5.1} & 78.2\textsubscript{ +3.8} & 62.6\textsubscript{ +7.4} \\
\hspace{1em}w/o CoT training & 57.6\textsubscript{ +2.8} & 67.1\textsubscript{ +4.6} & 63.4\textsubscript{ -3.2} & 49.6\textsubscript{ +3.1} & 37.0\textsubscript{ +12.8} & 58.0\textsubscript{ +0.7} & 71.4\textsubscript{ -3.0} & 57.7\textsubscript{ +2.5} \\
\hspace{1em}w/o training & 59.2\textsubscript{ +4.4} & 67.7\textsubscript{ +5.2} & 64.6\textsubscript{ -2.0} & 50.1\textsubscript{ +3.6} & 32.2\textsubscript{ +8.0} & 58.0\textsubscript{ +0.7} & 70.6\textsubscript{ -3.8} & 57.5\textsubscript{ +2.3} \\

\bottomrule
\end{tabular}
}
\end{table*}

\subsection{Experimental Setup}

\paragraph{Datasets and Metrics.}
In this paper, we evaluate the \model on the following tasks:
\begin{itemize}[left=0pt,noitemsep]
    \item Mathematical reasoning benchmarks, including GSM8k~\cite{cobbe2021training} and MATH500~\cite{lightman2024lets}. GSM8k is a widely used dataset for grade school math problems, while MATH500 is a subset extracted from the MATH dataset~\cite{hendrycksmath2021}, containing a diverse variety of challenging high school competition questions.
    \item Table question answering (TableQA) benchmarks, including WikiTQ~\cite{pasupat2015compositional} and FeTaQA~\cite{nan2022fetaqa}, which are two widely used TableQA benchmarks.
\end{itemize}
We focus on these tasks because both require substantial reasoning capabilities, and current LLMs still exhibit significant room for improvement in these domains. For the TableQA datasets, which require preprocessing to ensure format compatibility, we directly utilized the processed WikiTQ and FeTaQA test sets released by TableLLM~\cite{zhang2024tablellm}. Since our model is fine-tuned to adhere to a specific format, we extract answers that align with these format requirements from the LLM’s outputs to perform an exact match (EM) evaluation against the gold answers. Therefore, on the GSM8k, MATH500, and WikiTQ datasets, we use EM to calculate accuracy. Since the gold answers in the FeTaQA dataset are in the form of complete sentences, unlike the short answer phrases in WikiTQ, we select Rouge-L as the evaluation metric to better assess the quality of textual answers.

\paragraph{Baselines and Implementation.}
We evaluate our method on the following models: API-based models (GPT-4o~\cite{achiam2023gpt}, GLM-4-plus~\cite{glm2024chatglm}) and open-source models (Llama3-8B-Instruct~\cite{dubey2024llama}, Llama3.1-(8B,70B)-Instruct~\cite{dubey2024llama}, Qwen2-7B-Instruct~\cite{yang2024qwen2}, Qwen2.5-14B-Instruct~\cite{yang2024qwen2}). 

The baselines for comparison are as follows:
\begin{itemize}[left=0pt,noitemsep]
    \item CoT-prompting~\cite{wei2022chain}: We prompt the policy model to directly generate a single response without any post-processing techniques.
    \item Consistency-based methods:
    Self-consistency (SC)~\cite{wang2022self} and Universal Self-consistency (USC)~\cite{chen2023universal}. We use SC only on GSM8k and MATH500 since their answers are easy to extract, allowing majority voting based on exact match. Additionally, we adopt USC across all benchmarks.
    \item Best-of-N methods: We first employ the ArmoRM~\cite{wang2024interpretable} built on Llama3-8B, which is the top-ranked Llama3-8B-based model on RewardBench~\cite{lambert2024rewardbench}. To ensure a fairer comparison, we also train task-specific scalar reward model (Scalar RM) for MATH500 and WikiTQ datasets, which is the same as that used in our \model training, following the method in~\cite{cobbe2021training}.
    \item Synthesis methods: LMCOR~\cite{vernikos2023small}, which directly uses the gold answers $y_g$ from the benchmarks as the target responses for training its synthesis model, without incorporating candidate response analysis and integration through CoT process.
\end{itemize}

As described in Section~\ref{sub_sec:Training_data_preparation}, we utilize Llama3-8B-Instruct as the Sampling LLM to generate candidate responses and Llama3.1-70B-Instruct as the Response LLM to generate synthetic answers.
Our data generation pipeline expands the original 12k MATH training samples to 295k and the 18k WikiTQ training samples (from TableLLM) to 87k. 
During data filtering, for mathematical reasoning tasks, we filter out incorrect answers via exact matching with the gold answer. 
For TableQA, we follow TableLLM, using CritiqueLLM~\cite{ke2023critiquellm} to score the synthesized answers and retain only high-quality responses, as the generated answers are often lengthy and challenging to precisely match. Subsequently, we train the CoT-based \model on Llama3-8B-Instruct, as it demonstrates robust general capabilities and has not yet saturated the current datasets.

For all methods, we use Llama3.1-70B-Instruct as the backbone for tuning-free methods, and Llama3-8B-Instruct for training-based methods. Responses are generated by sampling k = 5 times from the policy model with T = 0.9 and Top-P = 0.9. To ensure reliability, each experiment is conducted three times, and the reported results are the average of these runs. Further details regarding experimental settings and implementation are provided in Appendix~\ref{app:baseline}.

\subsection{Main Results}

Table~\ref{tab:Main results} presents the experimental results of across four benchmarks. \textbf{The results demonstrate that our method significantly improves the reasoning performance of various policy models.} Below, we summarize the key findings:

\textbf{Our method consistently outperforms baselines across benchmarks.}
Across the four benchmarks, SC, ArmoRM, Scalar RM and our method generally improve the CoT outputs of various policy models. We evaluate a total of seven policy models. 1) On the MATH500 and WikiTQ datasets, our method (including both Llama3.1-70B-Instruct and \model-8B) outperforms all baseline methods on at least six policy models, achieving an average performance that is 2\% higher than the best-performing baseline. 
2) \model-8B demonstrates substantial improvements on larger models(e.g., GLM-4-Plus and Llama3.1-70B), outperforming all methods, even surpassing directly using Llama3.1-70B as \model.

\textbf{\model-8B generalizes well to unseen datasets}, even when the training data does not include examples from specific benchmarks. For example, despite the absence of GSM8k and FeTaQA datasets in the training data, \model-8B achieves superior performance on these tasks, highlighting its effectiveness in handling unseen data.

\textbf{\model-8B demonstrates robustness across structurally distinct models.}
\model-8B relies solely on Llama3-8B-Instruct to generate candidate responses and uses Llama3.1-70B-Instruct for generating final answers. Despite this, our method still shows significant improvements across a wide range of policy models, including GLM-4-Plus, GPT-4o, Qwen2-7B, and Qwen2.5-14B, which are structurally and behaviorally distinct from the models used during training.

These findings highlight that our method effectively learns a generalized reasoning and synthesis strategy, enabling it to enhance the outputs of diverse policy models without overfitting to the specific characteristics of the models used for training data generation.

\subsection{Ablation Studies}

We conduct ablation studies to evaluate the impact of the process of CoT analysis of candidate responses and training on data generation pipeline.

\textbf{Omitting CoT training affects robustness and generalization.}
In contrast to the original approach, where CoT plays a critical role, we directly utilize the Llama3.1-70B as the \model to provide answers to each question, generating high-quality responses. These responses are subsequently used as labels in the training data, thus skipping the CoT analysis of candidate responses. Table~\ref{tab:ablation results} indicates that skipping CoT training slightly improves average performance over CoT-prompting but remains inferior to \model-8B. Notably, for models like Llama3.1-70B and Qwen2.5-14B, the performance of skipping CoT training is even worse than the CoT-prompting, while on Llama3-8B testset it surpasses the \model-8B. This suggests that removing CoT can lead to overfitting to the Sampling LLM, thereby reducing generalization across different policy models. 

\textbf{Omitting training on generation data impacts reasoning and synthesis capabilities.}
To evaluate the effectiveness of our data generation pipeline, we directly utilized an untrained Llama3-8B-Instruct to regenerate candidate responses from various policy models on mathematical reasoning tasks. Table~\ref{tab:ablation results} illustrates that while the untrained model shows some synthesis improvements compared to CoT-prompting on MATH500 for some models (e.g., +8\% on Llama3-8B, +3.6\% on Llama3.1-8B), it underperforms for stronger models like Llama3.1-70B and Qwen2.5-14B. Similarly, on the GSM8k dataset, Llama3-8B fails to enhance most policy models, indicating that it still makes errors during reasoning and synthesis.

\subsection{Scaling Experiments}

We experiment with training data size and inference iterations to assess the scalability of our method. 

\paragraph{Training Data Scaling.}

\begin{figure}[t]
  \centering  \includegraphics[width=0.49\textwidth]{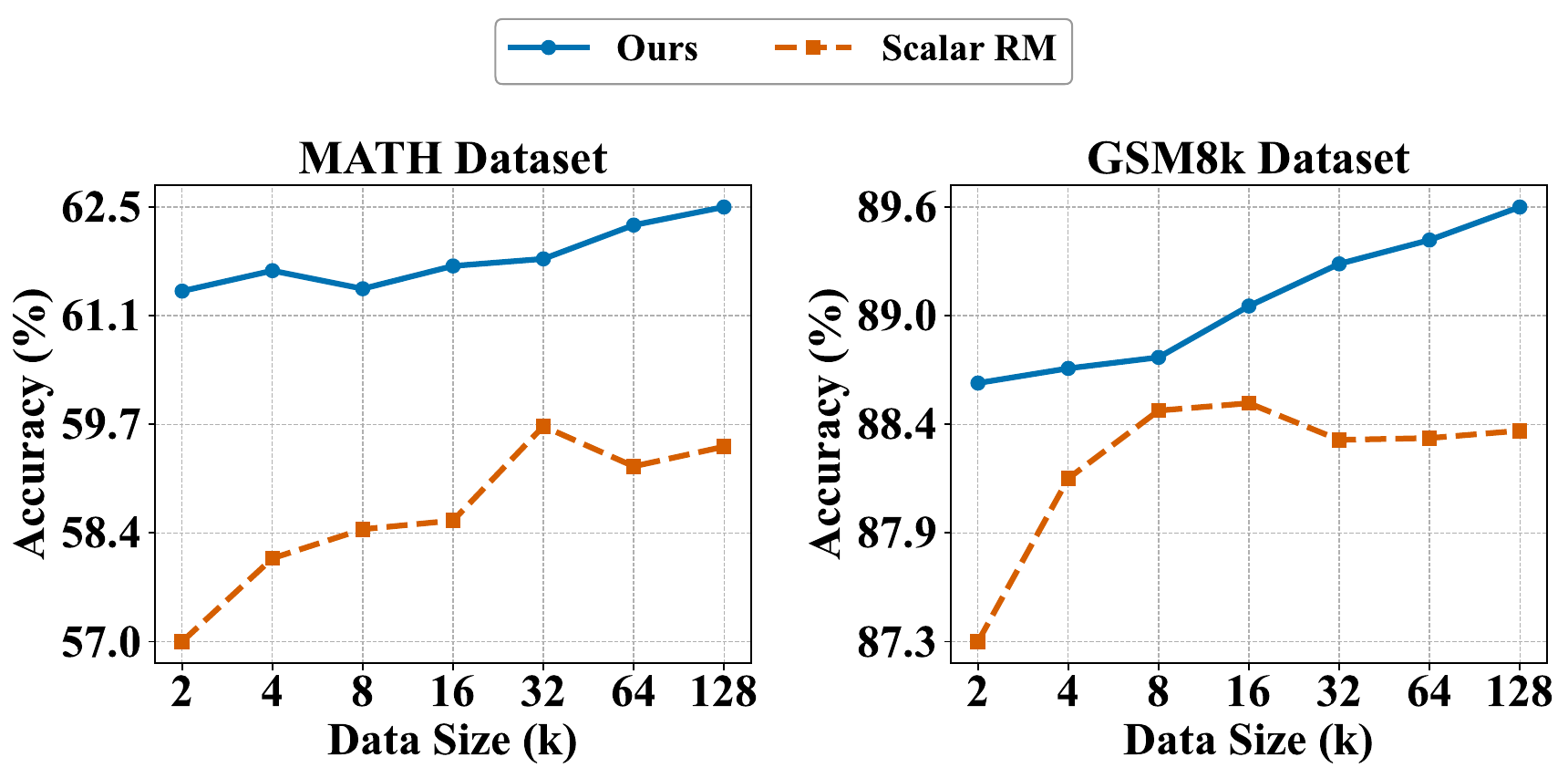}
  \caption{Effects of training data size on the performance of Scalar RM and Ours across the MATH500 and GSM8k datasets. As the data size increases, Scalar RM initially improves but subsequently declines while our method exhibits a consistent increase.}
  \label{fig:training scale}
\end{figure}

We evaluate the impact of training data size on both Scalar RM and our method. Specifically, we train the mathematical reasoning tasks with varying amounts of training data: 2k, 4k, 8k, 16k, 32k, 64k, 128k. 
These subsets are sampled separately from the training datasets of Scalar RM and our method in the main experiment, while keeping all other settings consistent. 
Figure~\ref{fig:training scale} illustrates the average performance trends of these two methods on the mathematical tasks as the data size increases.

\textbf{For our method, performance improves in a log-linear relationship with the data size} and consistently outperforms Scalar RM across all settings. Due to the limitations of the available data, training is terminated at 128k data points.

In contrast, Scalar RM exhibits a different trend: its performance initially improves with increasing data size but subsequently declines. This decline may be attributed to the limited diversity of training data, which is generated through multiple samplings of the original benchmark dataset, leading to a significant number of duplicate instructions.
Since Scalar RM is trained using positive and negative pairs and produces scalar scores during inference, it is more prone to ``overfitting'' on these repeated instructions.
Our generative approach, however, mitigates this issue by sampling multiple candidate answers for each question and synthesizing the final answer using CoT. Even when encountering repeated instructions, the sampling process ensures diverse candidate answers and reasoning chains, effectively transforming potentially repetitive training data into more diverse learning signals. This enables our approach more resilient to the lack of diversity in training data.

\paragraph{Inference Scaling.}

We further investigate the impact of the number of candidate responses generated by the policy models on the performance of SC, ArmoRM, Scalar RM, and our method. Using Llama3-8B and Qwen2-7B as policy models, we conduct 125 inferences on the MATH500 test set, with 1, 5, and 25 samples randomly selected from the candidate answers for comparison.

Due to input length constraints, our method cannot process all candidate responses simultaneously when their number is large. To address this, candidates are grouped into sets of five, with any remaining answers treated as a separate group. The synthesized outputs from each group are then further combined iteratively across groups.

\begin{figure}[t]
  \centering  \includegraphics[width=0.49\textwidth]{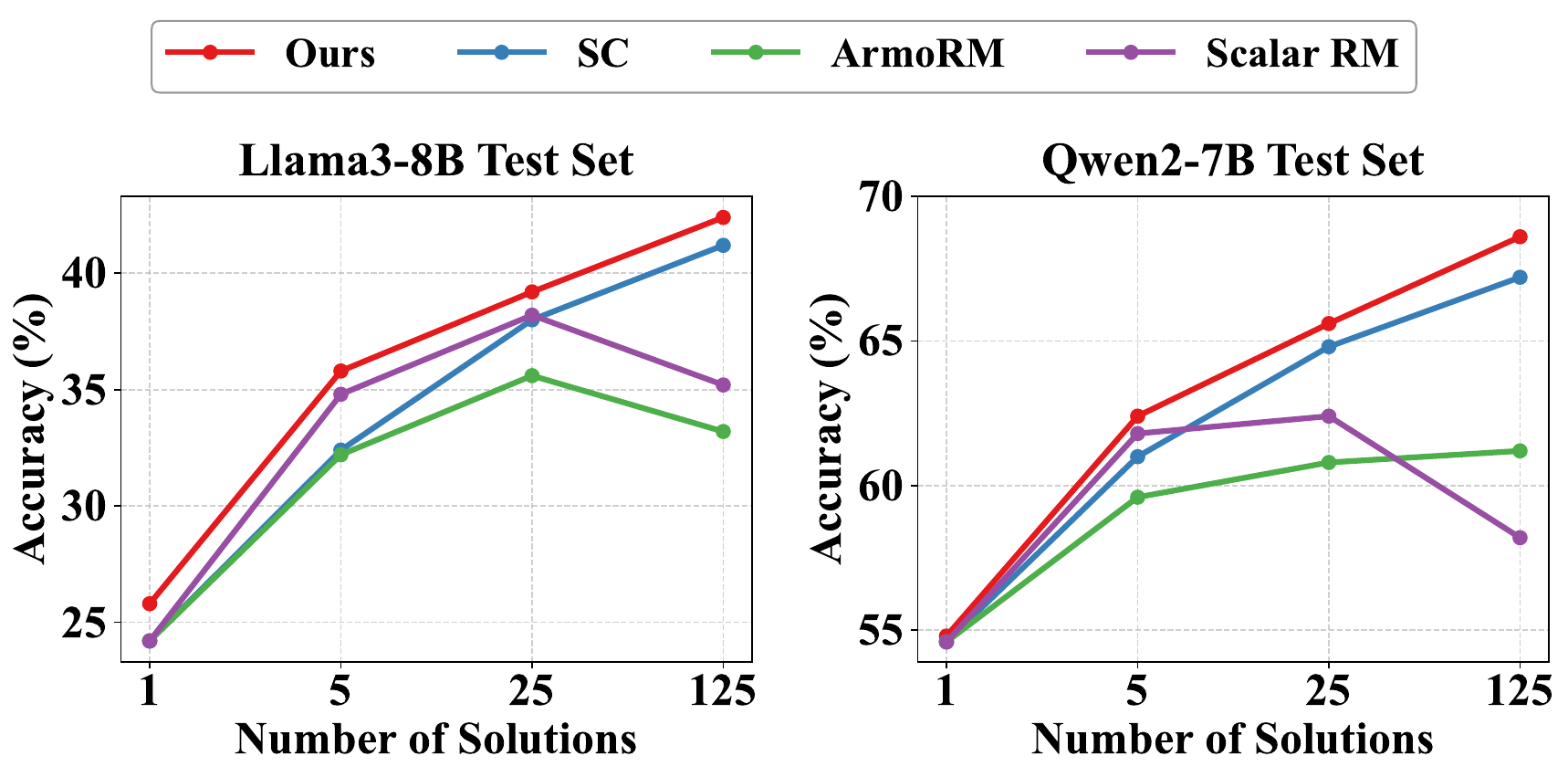}
  \caption{Effects of inference scaling on the performance of Scalar RM and Ours across the MATH500 and GSM8k datasets.}
  \label{fig:Inference Scaling}
\end{figure}

As shown in the Figure~\ref{fig:Inference Scaling}, \textbf{the performance of our method and SC improves steadily with more candidate responses}, indicating their robustness. In contrast, the performance of the ArmoRM and Scalar RM, both of which adopt Best-of-N methods, initially improve but then decline as candidate numbers grow. This decline can be attributed to the verifier being distracted by other solutions containing erroneous information while searching for the optimal solution, leading to a ``reward hacking'' phenomenon, which is consistent with the description in \cite{cobbe2021training}. Additionally, regarding cost-benefit trade-offs, our method achieves comparable results with only 5 candidates, while SC requires over 10, demonstrating that SC needs significantly more candidates to match our method's effectiveness. This underscores the efficiency and performance advantages of our method.

\subsection{Analysis}

We select Llama3-8B as the policy model and conduct a comparative analysis of SC, ArmoRM, Scalar RM and our method on the MATH500 test set. For each sample, the policy model generates 5 candidate responses, and we compare the number of correct answers as well as the final results after applying the post-processing methods, as shown in Table~\ref{tab:case_study}. Notably, our method can synthesize a correct answer even when all candidate responses are incorrect. Figure~\ref{fig:case_study} provides a detailed example.

\begin{table}[h]
  \setlength{\tabcolsep}{4pt}
  \begin{center}
    \scalebox{0.9}{
    \begin{tabular}{l|cccccc|c}
    \toprule
    Correct Count & 0 & 1 & 2 & 3 & 4 & 5 & Sum \\ 
    \midrule
    SC & 0 & 17 & 37 & \textbf{36} & \textbf{32} & \textbf{40} & 162\\ 
    ArmoRM & 0 & \textbf{36} & 35 & 22 & 31 & \textbf{40} & 164\\ 
    Scalar RM & 0 & \textbf{36} & 37 & 28 & 30 & \textbf{40} & 171\\ 
    \model-8B & \textbf{9} & 33 & \textbf{38} & \textbf{36} & \textbf{32} & \textbf{40} & \textbf{188}\\
    \bottomrule
    \end{tabular}}
    \caption{Correct answer distribution for different post-processing methods on Llama3-8B across 5 inferences.}
    \label{tab:case_study}
    \end{center}
\end{table}

\section{Conclusion}
\label{sec:bibtex}
This paper proposes CoT-based \model, a novel inference scaling strategy that integrates information from multiple candidate responses, enabling accurate answers. To enable a lightweight and cost-effective implementation, we introduce an automatic data generation pipeline, allowing smaller models to learn effectively from diverse data generated by large LLMs. Experiments across benchmarks demonstrate that CoT-based \model outperforms traditional methods like Best-of-N and Self-consistency.


\section*{Limitations}
\paragraph{Grouping and Synthesis of Candidate Responses.}
While our method supports the synthesis of multiple candidate answers, the input length limitation of the backbone model necessitates grouping the candidate answers for separate synthesis, followed by iterative synthesis of the grouped outputs. Although this approach leverages relationships among different candidate answers within the same group, it weakens the connections between candidates across different groups, and it also increases the overall inference cost. Nevertheless, despite the additional inference overhead introduced by the input length constraint, the lightweight Synthesizer model remains more efficient and computationally economical than the policy model, which requires multiple inference steps. With the emergence of more advanced long-context LLMs, our method is expected to achieve more efficient and higher-quality solution generation by leveraging multiple candidate answers.

\paragraph{Inference Overhead.}
Although our method processes multiple candidate answers simultaneously and performs synthesis in a single step, thereby significantly reducing the number of inference passes compared to the Best-of-N approach, which involves scoring and filtering each candidate answer, it still requires all candidate answers to be generated before synthesis can take place. This design introduces additional inference time overhead.

\section*{Acknowledgments}
This work is supported by the National Key Research \& Develop Plan (2023YFF0725100) and the National Natural Science Foundation of China (62322214, U23A20299, U24B20144, 62172424, 62276270).

\bibliography{aclbib}

\begin{thebibliography}{44}
\providecommand{\natexlab}[1]{#1}

\bibitem[{Achiam et~al.(2023)Achiam, Adler, Agarwal, Ahmad, Akkaya, Aleman, Almeida, Altenschmidt, Altman, Anadkat et~al.}]{achiam2023gpt}
Josh Achiam, Steven Adler, Sandhini Agarwal, Lama Ahmad, Ilge Akkaya, Florencia~Leoni Aleman, Diogo Almeida, Janko Altenschmidt, Sam Altman, Shyamal Anadkat, et~al. 2023.
\newblock Gpt-4 technical report.
\newblock \emph{arXiv preprint arXiv:2303.08774}.

\bibitem[{Brown et~al.(2024)Brown, Juravsky, Ehrlich, Clark, Le, R{\'e}, and Mirhoseini}]{brown2024large}
Bradley Brown, Jordan Juravsky, Ryan Ehrlich, Ronald Clark, Quoc~V Le, Christopher R{\'e}, and Azalia Mirhoseini. 2024.
\newblock Large language monkeys: Scaling inference compute with repeated sampling.
\newblock \emph{arXiv preprint arXiv:2407.21787}.

\bibitem[{Brown et~al.(2020)Brown, Mann, Ryder, Subbiah, Kaplan, Dhariwal, Neelakantan, Shyam, Sastry, Askell, Agarwal, Herbert-Voss, Krueger, Henighan, Child, Ramesh, Ziegler, Wu, Winter, Hesse, Chen, Sigler, Litwin, Gray, Chess, Clark, Berner, McCandlish, Radford, Sutskever, and Amodei}]{NEURIPS2020_1457c0d6}
Tom Brown, Benjamin Mann, Nick Ryder, Melanie Subbiah, Jared~D Kaplan, Prafulla Dhariwal, Arvind Neelakantan, Pranav Shyam, Girish Sastry, Amanda Askell, Sandhini Agarwal, Ariel Herbert-Voss, Gretchen Krueger, Tom Henighan, Rewon Child, Aditya Ramesh, Daniel Ziegler, Jeffrey Wu, Clemens Winter, Chris Hesse, Mark Chen, Eric Sigler, Mateusz Litwin, Scott Gray, Benjamin Chess, Jack Clark, Christopher Berner, Sam McCandlish, Alec Radford, Ilya Sutskever, and Dario Amodei. 2020.
\newblock \href {https://proceedings.neurips.cc/paper_files/paper/2020/file/1457c0d6bfcb4967418bfb8ac142f64a-Paper.pdf} {Language models are few-shot learners}.
\newblock In \emph{Advances in Neural Information Processing Systems}, volume~33, pages 1877--1901. Curran Associates, Inc.

\bibitem[{Chen et~al.(2023)Chen, Aksitov, Alon, Ren, Xiao, Yin, Prakash, Sutton, Wang, and Zhou}]{chen2023universal}
Xinyun Chen, Renat Aksitov, Uri Alon, Jie Ren, Kefan Xiao, Pengcheng Yin, Sushant Prakash, Charles Sutton, Xuezhi Wang, and Denny Zhou. 2023.
\newblock Universal self-consistency for large language model generation.
\newblock \emph{arXiv preprint arXiv:2311.17311}.

\bibitem[{Chowdhery et~al.(2023)Chowdhery, Narang, Devlin, Bosma, Mishra, Roberts, Barham, Chung, Sutton, Gehrmann, Schuh, Shi, Tsvyashchenko, Maynez, Rao, Barnes, Tay, Shazeer, Prabhakaran, Reif, Du, Hutchinson, Pope, Bradbury, Austin, Isard, Gur-Ari, Yin, Duke, Levskaya, Ghemawat, Dev, Michalewski, Garcia, Misra, Robinson, Fedus, Zhou, Ippolito, Luan, Lim, Zoph, Spiridonov, Sepassi, Dohan, Agrawal, Omernick, Dai, Pillai, Pellat, Lewkowycz, Moreira, Child, Polozov, Lee, Zhou, Wang, Saeta, Diaz, Firat, Catasta, Wei, Meier-Hellstern, Eck, Dean, Petrov, and Fiedel}]{Aakanksha2023palm}
Aakanksha Chowdhery, Sharan Narang, Jacob Devlin, Maarten Bosma, Gaurav Mishra, Adam Roberts, Paul Barham, Hyung~Won Chung, Charles Sutton, Sebastian Gehrmann, Parker Schuh, Kensen Shi, Sasha Tsvyashchenko, Joshua Maynez, Abhishek Rao, Parker Barnes, Yi~Tay, Noam Shazeer, Vinodkumar Prabhakaran, Emily Reif, Nan Du, Ben Hutchinson, Reiner Pope, James Bradbury, Jacob Austin, Michael Isard, Guy Gur-Ari, Pengcheng Yin, Toju Duke, Anselm Levskaya, Sanjay Ghemawat, Sunipa Dev, Henryk Michalewski, Xavier Garcia, Vedant Misra, Kevin Robinson, Liam Fedus, Denny Zhou, Daphne Ippolito, David Luan, Hyeontaek Lim, Barret Zoph, Alexander Spiridonov, Ryan Sepassi, David Dohan, Shivani Agrawal, Mark Omernick, Andrew~M. Dai, Thanumalayan~Sankaranarayana Pillai, Marie Pellat, Aitor Lewkowycz, Erica Moreira, Rewon Child, Oleksandr Polozov, Katherine Lee, Zongwei Zhou, Xuezhi Wang, Brennan Saeta, Mark Diaz, Orhan Firat, Michele Catasta, Jason Wei, Kathy Meier-Hellstern, Douglas Eck, Jeff Dean, Slav Petrov, and Noah Fiedel. 2023.
\newblock \href {http://jmlr.org/papers/v24/22-1144.html} {Palm: Scaling language modeling with pathways}.
\newblock \emph{Journal of Machine Learning Research}, 24(240):1--113.

\bibitem[{Christiano et~al.(2017)Christiano, Leike, Brown, Martic, Legg, and Amodei}]{christiano2017deep}
Paul~F Christiano, Jan Leike, Tom Brown, Miljan Martic, Shane Legg, and Dario Amodei. 2017.
\newblock Deep reinforcement learning from human preferences.
\newblock \emph{Advances in neural information processing systems}, 30.

\bibitem[{Cobbe et~al.(2021)Cobbe, Kosaraju, Bavarian, Chen, Jun, Kaiser, Plappert, Tworek, Hilton, Nakano et~al.}]{cobbe2021training}
Karl Cobbe, Vineet Kosaraju, Mohammad Bavarian, Mark Chen, Heewoo Jun, Lukasz Kaiser, Matthias Plappert, Jerry Tworek, Jacob Hilton, Reiichiro Nakano, et~al. 2021.
\newblock Training verifiers to solve math word problems.
\newblock \emph{arXiv preprint arXiv:2110.14168}.

\bibitem[{Dubey et~al.(2024)Dubey, Jauhri, Pandey, Kadian, Al-Dahle, Letman, Mathur, Schelten, Yang, Fan et~al.}]{dubey2024llama}
Abhimanyu Dubey, Abhinav Jauhri, Abhinav Pandey, Abhishek Kadian, Ahmad Al-Dahle, Aiesha Letman, Akhil Mathur, Alan Schelten, Amy Yang, Angela Fan, et~al. 2024.
\newblock The llama 3 herd of models.
\newblock \emph{arXiv preprint arXiv:2407.21783}.

\bibitem[{Farinhas et~al.(2023)Farinhas, de~Souza, and Martins}]{farinhas2023empirical}
Ant{\'o}nio Farinhas, Jos{\'e}~GC de~Souza, and Andr{\'e}~FT Martins. 2023.
\newblock An empirical study of translation hypothesis ensembling with large language models.
\newblock \emph{arXiv preprint arXiv:2310.11430}.

\bibitem[{GLM et~al.(2024)GLM, Zeng, Xu, Wang, Zhang, Yin, Zhang, Rojas, Feng, Zhao et~al.}]{glm2024chatglm}
Team GLM, Aohan Zeng, Bin Xu, Bowen Wang, Chenhui Zhang, Da~Yin, Dan Zhang, Diego Rojas, Guanyu Feng, Hanlin Zhao, et~al. 2024.
\newblock Chatglm: A family of large language models from glm-130b to glm-4 all tools.
\newblock \emph{arXiv preprint arXiv:2406.12793}.

\bibitem[{He et~al.(2024)He, Luo, Bai, Hu, Thai, Shen, Hu, Han, Huang, Zhang et~al.}]{he2024olympiadbench}
Chaoqun He, Renjie Luo, Yuzhuo Bai, Shengding Hu, Zhen~Leng Thai, Junhao Shen, Jinyi Hu, Xu~Han, Yujie Huang, Yuxiang Zhang, et~al. 2024.
\newblock Olympiadbench: A challenging benchmark for promoting agi with olympiad-level bilingual multimodal scientific problems.
\newblock \emph{arXiv preprint arXiv:2402.14008}.

\bibitem[{Hendrycks et~al.(2021)Hendrycks, Burns, Kadavath, Arora, Basart, Tang, Song, and Steinhardt}]{hendrycksmath2021}
Dan Hendrycks, Collin Burns, Saurav Kadavath, Akul Arora, Steven Basart, Eric Tang, Dawn Song, and Jacob Steinhardt. 2021.
\newblock Measuring mathematical problem solving with the math dataset.
\newblock \emph{NeurIPS}.

\bibitem[{Holtzman et~al.(2020)Holtzman, Buys, Du, Forbes, and Choi}]{Holtzman2020The}
Ari Holtzman, Jan Buys, Li~Du, Maxwell Forbes, and Yejin Choi. 2020.
\newblock \href {https://openreview.net/forum?id=rygGQyrFvH} {The curious case of neural text degeneration}.
\newblock In \emph{International Conference on Learning Representations}.

\bibitem[{Jiang et~al.(2023)Jiang, Ren, and Lin}]{jiang2023llm}
Dongfu Jiang, Xiang Ren, and Bill~Yuchen Lin. 2023.
\newblock Llm-blender: Ensembling large language models with pairwise ranking and generative fusion.
\newblock \emph{arXiv preprint arXiv:2306.02561}.

\bibitem[{Ke et~al.(2023)Ke, Wen, Feng, Liu, Lei, Cheng, Wang, Zeng, Dong, Wang et~al.}]{ke2023critiquellm}
Pei Ke, Bosi Wen, Zhuoer Feng, Xiao Liu, Xuanyu Lei, Jiale Cheng, Shengyuan Wang, Aohan Zeng, Yuxiao Dong, Hongning Wang, et~al. 2023.
\newblock Critiquellm: Scaling llm-as-critic for effective and explainable evaluation of large language model generation.
\newblock \emph{arXiv preprint arXiv:2311.18702}.

\bibitem[{Kwon et~al.(2023)Kwon, Li, Zhuang, Sheng, Zheng, Yu, Gonzalez, Zhang, and Stoica}]{kwon2023efficient}
Woosuk Kwon, Zhuohan Li, Siyuan Zhuang, Ying Sheng, Lianmin Zheng, Cody~Hao Yu, Joseph Gonzalez, Hao Zhang, and Ion Stoica. 2023.
\newblock Efficient memory management for large language model serving with pagedattention.
\newblock In \emph{Proceedings of the 29th Symposium on Operating Systems Principles}, pages 611--626.

\bibitem[{Lambert et~al.(2024)Lambert, Pyatkin, Morrison, Miranda, Lin, Chandu, Dziri, Kumar, Zick, Choi et~al.}]{lambert2024rewardbench}
Nathan Lambert, Valentina Pyatkin, Jacob Morrison, LJ~Miranda, Bill~Yuchen Lin, Khyathi Chandu, Nouha Dziri, Sachin Kumar, Tom Zick, Yejin Choi, et~al. 2024.
\newblock Rewardbench: Evaluating reward models for language modeling.
\newblock \emph{arXiv preprint arXiv:2403.13787}.

\bibitem[{Li et~al.(2022)Li, Choi, Chung, Kushman, Schrittwieser, Leblond, Eccles, Keeling, Gimeno, Dal~Lago et~al.}]{li2022competition}
Yujia Li, David Choi, Junyoung Chung, Nate Kushman, Julian Schrittwieser, R{\'e}mi Leblond, Tom Eccles, James Keeling, Felix Gimeno, Agustin Dal~Lago, et~al. 2022.
\newblock Competition-level code generation with alphacode.
\newblock \emph{Science}, 378(6624):1092--1097.

\bibitem[{Lightman et~al.(2024)Lightman, Kosaraju, Burda, Edwards, Baker, Lee, Leike, Schulman, Sutskever, and Cobbe}]{lightman2024lets}
Hunter Lightman, Vineet Kosaraju, Yuri Burda, Harrison Edwards, Bowen Baker, Teddy Lee, Jan Leike, John Schulman, Ilya Sutskever, and Karl Cobbe. 2024.
\newblock \href {https://openreview.net/forum?id=v8L0pN6EOi} {Let's verify step by step}.
\newblock In \emph{The Twelfth International Conference on Learning Representations}.

\bibitem[{Loshchilov(2017)}]{loshchilov2017decoupled}
I~Loshchilov. 2017.
\newblock Decoupled weight decay regularization.
\newblock \emph{arXiv preprint arXiv:1711.05101}.

\bibitem[{Nakano et~al.(2021)Nakano, Hilton, Balaji, Wu, Ouyang, Kim, Hesse, Jain, Kosaraju, Saunders et~al.}]{nakano2021webgpt}
Reiichiro Nakano, Jacob Hilton, Suchir Balaji, Jeff Wu, Long Ouyang, Christina Kim, Christopher Hesse, Shantanu Jain, Vineet Kosaraju, William Saunders, et~al. 2021.
\newblock Webgpt: Browser-assisted question-answering with human feedback.
\newblock \emph{arXiv preprint arXiv:2112.09332}.

\bibitem[{Nan et~al.(2022)Nan, Hsieh, Mao, Lin, Verma, Zhang, Kry{\'s}ci{\'n}ski, Schoelkopf, Kong, Tang et~al.}]{nan2022fetaqa}
Linyong Nan, Chiachun Hsieh, Ziming Mao, Xi~Victoria Lin, Neha Verma, Rui Zhang, Wojciech Kry{\'s}ci{\'n}ski, Hailey Schoelkopf, Riley Kong, Xiangru Tang, et~al. 2022.
\newblock Fetaqa: Free-form table question answering.
\newblock \emph{Transactions of the Association for Computational Linguistics}, 10:35--49.

\bibitem[{Nye et~al.(2021)Nye, Andreassen, Gur-Ari, Michalewski, Austin, Bieber, Dohan, Lewkowycz, Bosma, Luan et~al.}]{nye2021show}
Maxwell Nye, Anders~Johan Andreassen, Guy Gur-Ari, Henryk Michalewski, Jacob Austin, David Bieber, David Dohan, Aitor Lewkowycz, Maarten Bosma, David Luan, et~al. 2021.
\newblock Show your work: Scratchpads for intermediate computation with language models.
\newblock \emph{arXiv preprint arXiv:2112.00114}.

\bibitem[{Ouyang et~al.(2022)Ouyang, Wu, Jiang, Almeida, Wainwright, Mishkin, Zhang, Agarwal, Slama, Ray et~al.}]{ouyang2022training}
Long Ouyang, Jeffrey Wu, Xu~Jiang, Diogo Almeida, Carroll Wainwright, Pamela Mishkin, Chong Zhang, Sandhini Agarwal, Katarina Slama, Alex Ray, et~al. 2022.
\newblock Training language models to follow instructions with human feedback.
\newblock \emph{Advances in neural information processing systems}, 35:27730--27744.

\bibitem[{Pasupat and Liang(2015)}]{pasupat2015compositional}
Panupong Pasupat and Percy Liang. 2015.
\newblock Compositional semantic parsing on semi-structured tables.
\newblock \emph{arXiv preprint arXiv:1508.00305}.

\bibitem[{Rae et~al.(2021)Rae, Borgeaud, Cai, Millican, Hoffmann, Song, Aslanides, Henderson, Ring, Young et~al.}]{rae2021scaling}
Jack~W Rae, Sebastian Borgeaud, Trevor Cai, Katie Millican, Jordan Hoffmann, Francis Song, John Aslanides, Sarah Henderson, Roman Ring, Susannah Young, et~al. 2021.
\newblock Scaling language models: Methods, analysis \& insights from training gopher.
\newblock \emph{arXiv preprint arXiv:2112.11446}.

\bibitem[{Shi et~al.(2022)Shi, Fried, Ghazvininejad, Zettlemoyer, and Wang}]{shi2022natural}
Freda Shi, Daniel Fried, Marjan Ghazvininejad, Luke Zettlemoyer, and Sida~I Wang. 2022.
\newblock Natural language to code translation with execution.
\newblock In \emph{Proceedings of the 2022 Conference on Empirical Methods in Natural Language Processing}, pages 3533--3546.

\bibitem[{Singhal et~al.(2023)Singhal, Goyal, Xu, and Durrett}]{singhal2023long}
Prasann Singhal, Tanya Goyal, Jiacheng Xu, and Greg Durrett. 2023.
\newblock A long way to go: Investigating length correlations in rlhf.
\newblock \emph{arXiv preprint arXiv:2310.03716}.

\bibitem[{Skalse et~al.(2022)Skalse, Howe, Krasheninnikov, and Krueger}]{skalse2022defining}
Joar Skalse, Nikolaus Howe, Dmitrii Krasheninnikov, and David Krueger. 2022.
\newblock Defining and characterizing reward gaming.
\newblock \emph{Advances in Neural Information Processing Systems}, 35:9460--9471.

\bibitem[{Stiennon et~al.(2020)Stiennon, Ouyang, Wu, Ziegler, Lowe, Voss, Radford, Amodei, and Christiano}]{stiennon2020learning}
Nisan Stiennon, Long Ouyang, Jeffrey Wu, Daniel Ziegler, Ryan Lowe, Chelsea Voss, Alec Radford, Dario Amodei, and Paul~F Christiano. 2020.
\newblock Learning to summarize with human feedback.
\newblock \emph{Advances in Neural Information Processing Systems}, 33:3008--3021.

\bibitem[{Tong et~al.(2024)Tong, Zhang, Wang, Wu, and He}]{tong2024dart}
Yuxuan Tong, Xiwen Zhang, Rui Wang, Ruidong Wu, and Junxian He. 2024.
\newblock Dart-math: Difficulty-aware rejection tuning for mathematical problem-solving.
\newblock \emph{arXiv preprint arXiv:2407.13690}.

\bibitem[{Touvron et~al.(2023)Touvron, Martin, Stone, Albert, Almahairi, Babaei, Bashlykov, Batra, Bhargava, Bhosale et~al.}]{touvron2023llama}
Hugo Touvron, Louis Martin, Kevin Stone, Peter Albert, Amjad Almahairi, Yasmine Babaei, Nikolay Bashlykov, Soumya Batra, Prajjwal Bhargava, Shruti Bhosale, et~al. 2023.
\newblock Llama 2: Open foundation and fine-tuned chat models.
\newblock \emph{arXiv preprint arXiv:2307.09288}.

\bibitem[{Uesato et~al.(2022)Uesato, Kushman, Kumar, Song, Siegel, Wang, Creswell, Irving, and Higgins}]{uesato2022solving}
Jonathan Uesato, Nate Kushman, Ramana Kumar, Francis Song, Noah Siegel, Lisa Wang, Antonia Creswell, Geoffrey Irving, and Irina Higgins. 2022.
\newblock Solving math word problems with process-and outcome-based feedback.
\newblock \emph{arXiv preprint arXiv:2211.14275}.

\bibitem[{Vernikos et~al.(2023)Vernikos, Bra{\v{z}}inskas, Adamek, Mallinson, Severyn, and Malmi}]{vernikos2023small}
Giorgos Vernikos, Arthur Bra{\v{z}}inskas, Jakub Adamek, Jonathan Mallinson, Aliaksei Severyn, and Eric Malmi. 2023.
\newblock Small language models improve giants by rewriting their outputs.
\newblock \emph{arXiv preprint arXiv:2305.13514}.

\bibitem[{Vernikos and Popescu-Belis(2024)}]{vernikos2024don}
Giorgos Vernikos and Andrei Popescu-Belis. 2024.
\newblock Don't rank, combine! combining machine translation hypotheses using quality estimation.
\newblock \emph{arXiv preprint arXiv:2401.06688}.

\bibitem[{Wang et~al.(2024)Wang, Xiong, Xie, Zhao, and Zhang}]{wang2024interpretable}
Haoxiang Wang, Wei Xiong, Tengyang Xie, Han Zhao, and Tong Zhang. 2024.
\newblock Interpretable preferences via multi-objective reward modeling and mixture-of-experts.
\newblock \emph{arXiv preprint arXiv:2406.12845}.

\bibitem[{Wang et~al.(2022)Wang, Wei, Schuurmans, Le, Chi, Narang, Chowdhery, and Zhou}]{wang2022self}
Xuezhi Wang, Jason Wei, Dale Schuurmans, Quoc Le, Ed~Chi, Sharan Narang, Aakanksha Chowdhery, and Denny Zhou. 2022.
\newblock Self-consistency improves chain of thought reasoning in language models.
\newblock \emph{arXiv preprint arXiv:2203.11171}.

\bibitem[{Wei et~al.(2022)Wei, Wang, Schuurmans, Bosma, Xia, Chi, Le, Zhou et~al.}]{wei2022chain}
Jason Wei, Xuezhi Wang, Dale Schuurmans, Maarten Bosma, Fei Xia, Ed~Chi, Quoc~V Le, Denny Zhou, et~al. 2022.
\newblock Chain-of-thought prompting elicits reasoning in large language models.
\newblock \emph{Advances in neural information processing systems}, 35:24824--24837.

\bibitem[{Wu et~al.(2024)Wu, Sun, Li, Welleck, and Yang}]{wu2024empirical}
Yangzhen Wu, Zhiqing Sun, Shanda Li, Sean Welleck, and Yiming Yang. 2024.
\newblock An empirical analysis of compute-optimal inference for problem-solving with language models.
\newblock \emph{arXiv preprint arXiv:2408.00724}.

\bibitem[{Yang et~al.(2024)Yang, Yang, Hui, Zheng, Yu, Zhou, Li, Li, Liu, Huang et~al.}]{yang2024qwen2}
An~Yang, Baosong Yang, Binyuan Hui, Bo~Zheng, Bowen Yu, Chang Zhou, Chengpeng Li, Chengyuan Li, Dayiheng Liu, Fei Huang, et~al. 2024.
\newblock Qwen2 technical report.
\newblock \emph{arXiv preprint arXiv:2407.10671}.

\bibitem[{Yu et~al.(2023)Yu, Gao, and Wang}]{yu2023outcome}
Fei Yu, Anningzhe Gao, and Benyou Wang. 2023.
\newblock Outcome-supervised verifiers for planning in mathematical reasoning.
\newblock \emph{arXiv preprint arXiv:2311.09724}.

\bibitem[{Yu et~al.(2024)Yu, Chen, Zhang, Tan, Zhu, Pang, Qian, Wang, Gururangan, Zhang et~al.}]{yu2024self}
Yue Yu, Zhengxing Chen, Aston Zhang, Liang Tan, Chenguang Zhu, Richard~Yuanzhe Pang, Yundi Qian, Xuewei Wang, Suchin Gururangan, Chao Zhang, et~al. 2024.
\newblock Self-generated critiques boost reward modeling for language models.
\newblock \emph{arXiv preprint arXiv:2411.16646}.

\bibitem[{Zhang et~al.(2024{\natexlab{a}})Zhang, Hosseini, Bansal, Kazemi, Kumar, and Agarwal}]{zhang2024generative}
Lunjun Zhang, Arian Hosseini, Hritik Bansal, Mehran Kazemi, Aviral Kumar, and Rishabh Agarwal. 2024{\natexlab{a}}.
\newblock Generative verifiers: Reward modeling as next-token prediction.
\newblock \emph{arXiv preprint arXiv:2408.15240}.

\bibitem[{Zhang et~al.(2024{\natexlab{b}})Zhang, Zhang, Ma, Li, Zhang, Li, Yao, Xu, Zhou, Zhang-Li et~al.}]{zhang2024tablellm}
Xiaokang Zhang, Jing Zhang, Zeyao Ma, Yang Li, Bohan Zhang, Guanlin Li, Zijun Yao, Kangli Xu, Jinchang Zhou, Daniel Zhang-Li, et~al. 2024{\natexlab{b}}.
\newblock Tablellm: Enabling tabular data manipulation by llms in real office usage scenarios.
\newblock \emph{arXiv preprint arXiv:2403.19318}.

\end{thebibliography}

\appendix

\section{License}
Our research utilizes the training datasets from MATH and WikiTQ as foundational resources. These datasets are distributed under the Apache 2.0 license, which permits users to freely use, modify, reproduce, and share the software for both personal and commercial purposes.

In alignment with the principles of open access, we commit to publicly releasing the training data upon the acceptance of this work. The released data will be licensed under the CC BY-SA 4.0 license, ensuring its reuse and redistribution are permitted, provided that derivative works maintain the same licensing terms.

\section{Additional Ablation Experiments}
\label{app:ablation}
\begin{table*}[ht]
\centering
\caption{The impact of LLM Repair and Response LLM multiple sampling in ablation experiments of the data generation pipeline.}
\label{tab:ablation_appendix}
\resizebox{\textwidth}{!}{%
\begin{tabular}{lllllllll}
\toprule
\multirow{2}{*}{} & & \multicolumn{7}{c}{Policy Models} \\ 
\cmidrule{2-9}
Method & GLM4-api & GPT-4o & Llama3.1-70B & Llama3.1-8B & Llama3-8B & Qwen2-7B & Qwen2.5-14B & Average \\
\midrule
\grayline \multicolumn{9}{c}{\textbf{GSM8k}} \\
\midrule
\model-8B(ours) & 91.4 & 93.0 & 94.0 & 86.1 & 81.3 & 86.4 & 92.7 & 89.3 \\
\quad w/o LLM Repair & 90.6\textsubscript{ -0.8} & 92.9\textsubscript{ -0.1} & 93.6\textsubscript{ -0.4} & 85.7\textsubscript{ -0.4} & 81.5\textsubscript{ +0.2} & 86.6\textsubscript{ +0.2} & 92.0\textsubscript{ -0.7} & 89.0\textsubscript{ -0.3} \\
\quad w/o multiple responses & 90.5\textsubscript{ -0.9} & 91.8\textsubscript{ -1.2} & 93.4\textsubscript{ -0.6} & 84.6\textsubscript{ -1.5} & 80.8\textsubscript{ -0.5} & 85.4\textsubscript{ -1.0} & 91.8\textsubscript{ -0.9} & 88.3\textsubscript{ -1.0} \\
\midrule
\grayline \multicolumn{9}{c}{\textbf{MATH500}} \\
\midrule
\model-8B(ours) & 64.4 & 72.8 & 69.6 & 54.6 & 36.0 & 62.4 & 78.2 & 62.6 \\
\quad w/o LLM Repair & 63.6\textsubscript{ -0.8} & 72.3\textsubscript{ -0.5} & 68.4\textsubscript{ -1.2} & 53.6\textsubscript{ -1.0} & 35.2\textsubscript{ -0.8} & 60.8\textsubscript{ -1.6} & 77.8\textsubscript{ -0.4} & 61.7\textsubscript{ -0.9} \\
\quad w/o multiple responses & 63.4\textsubscript{ -1.0} & 71.7\textsubscript{ -1.1} & 68.0\textsubscript{ -1.6} & 53.4\textsubscript{ -1.2} & 33.6\textsubscript{ -2.4} & 60.6\textsubscript{ -1.8} & 77.0\textsubscript{ -1.2} & 61.1\textsubscript{ -1.5} \\
\bottomrule
\end{tabular}
}
\end{table*}

Table~\ref{tab:ablation_appendix} provides additional results on the impact of LLM Repair and Response LLM Sampling. The results show that both LLM Repair and Response LLM Sampling contribute positively to the overall performance of the \model-8B across different policy models and benchmarks. Removing LLM Repair leads to a slight drop in accuracy, while omitting Response LLM Sampling results in a more noticeable decline. This is because without Response LLM Sampling, the training data is limited to only 9k examples after filtering from the original 12k dataset. In contrast, removing LLM Repair still leaves 245k training examples, reducing the dataset by only 50k compared to the 295k used in the main experiment. This difference in dataset size explains the varying impact of the two ablation studies. Overall, incorporating both methods leads to more reliable and accurate responses.

\section{Experiment Settings}
\label{app:baseline}
\begin{table}[htbp]
  \setlength{\tabcolsep}{4pt}
  \begin{center}
    \scalebox{0.93}{
    \begin{tabular} {ll}
      \toprule
      Hyperparameters & Value \\
      \midrule
      Temperature & 0.9 \\
      Top P & 0.9 \\
      Max Tokens & 1024 \\
      Frequency Penalty & 0 \\
      Presence Penalty & 0 \\
      \bottomrule
      \end{tabular}}
    \caption{The hyperparameters of LLMs for candidate response generation.}
    \label{tab:parameters}
  \end{center}
\end{table}

\begin{table*}[ht]
\centering
  \caption{Information of baseline models.}
  \label{tab:model_info}
  \centering
  \resizebox{\linewidth}{!}{
  \begin{tabular}{l|l}
    \toprule
    Model & Website Url \\
    \midrule
    Llama3-8B-Instruct & \url{https://huggingface.co/meta-llama/Meta-Llama-3-8B-Instruct} \\
    Llama-3.1-8B-Instruct & \url{https://huggingface.co/meta-llama/Llama-3.1-8B-Instruct} \\
    Llama-3.1-70B-Instruct & \url{https://huggingface.co/meta-llama/Llama-3.1-70B-Instruct} \\
    Qwen2-7B-Instruct & \url{https://huggingface.co/Qwen/Qwen2-7B-Instruct} \\
    Qwen2.5-14B-Instruct & \url{https://huggingface.co/Qwen/Qwen2.5-14B-Instruct} \\
    ArmoRM-Llama3-8B-v0.1 & \url{https://huggingface.co/mistralai/Mixtral-8x7B-Instruct-v0.1} \\
    \midrule
    GPT-4o-2024-0513 & \url{https://platform.openai.com/overview} \\
    GLM-4-plus & \url{https://bigmodel.cn/} \\
    \bottomrule
  \end{tabular}}
\end{table*}

\begin{table}[h]
\setlength{\tabcolsep}{4pt}
  \begin{center}
    \scalebox{0.83}{
\begin{tabular}{lccc}
\toprule
Benchmark & Train Size & Extended Size & Test Size \\
\midrule
GSM8k  & -    & -     & 1319 \\
MATH   & 12k  & 295k  & 500  \\
FeTaQA & -    & -     & 753  \\
WikiTQ & 18k  & 87k   & 633  \\
\bottomrule
\end{tabular}}
\caption{Benchmark dataset sizes for train and test.}
\label{tab:benchmark_sizes}
\end{center}
\end{table}

\subsection{Experimental Environment}

Our experiments are conducted on a server running the Ubuntu 22.04 operating system, utilizing PyTorch version 2.4.0. The system is equipped with 8 NVIDIA A800 80GB GPUs, an Intel (R) Xeon (R) Platinum 8358 CPU, and 2048GB of RAM.

\subsection{Baselines and Implementation}
\label{app:baseline_details}
The policy models are divided into two types, including API-based models and open-source models. For the open-source models that require us to deploy them ourselves, we use \textit{PyTorch}, \textit{Transformers}, and \textit{vLLM}~\cite{kwon2023efficient} to load the models. For each policy model, we generate five candidate responses. The inference parameters can be referred to in Table ~\ref{tab:parameters}. For detailed information about each policy model, please refer to Table~\ref{tab:model_info}. 



Unlike traditional prompting methods that rely on iterative refinement, debate, or multi-step reasoning with decomposition and retrieval, our approach adopts an end-to-end synthesis strategy for efficiency and simplicity. Instead of introducing additional computational overhead, we directly generate a high-quality response in a single step by leveraging CoT reasoning to integrate information from multiple candidates. This streamlined design prioritizes efficiency, focusing on single-pass solution generation over multi-step reasoning.

To better evaluate the effectiveness of our approach, we further select the following baseline methods for comparison:
\begin{itemize}
    \item CoT-prompting: We use the prompts specifically designed for each dataset to generate CoT responses, with prompt details provided in Appendix~\ref{app:prompt}.
    CoT-prompting serves as a fundamental baseline, allowing us to assess the performance gains achieved by various post-processing methods.
    \item Consistency-based Methods: For mathematical reasoning, we adopt the standard Self-consistency configuration. However, since TableQA involves open-ended responses, it is challenging to extract precise answers for majority voting, making it unsuitable for Self-consistency tasks. Therefore, we also utilize a variant of consistency-based methods, USC, across four datasets. For a fair comparison, we use Llama3.1-70B-Instruct as the base model for USC experiments, enabling majority voting among candidate answers under consistent experimental conditions.
    \item Best-of-N Methods: We initially select ArmoRM-Llama3-8B-v0.1, which currently achieves the highest reasoning performance on the Reward Bench leaderboard based on Llama3-8B-Instruct. Additionally, we train a standard scalar reward model following the approach in ~\cite{cobbe2021training}. Specifically, for the MATH and WikiTQ datasets, to achieve a consistent training data size with our method, we use Llama3-8B-Instruct to generate 15 samples per data instance with a temperature of 0.9. Correct answers (including the gold answer) are treated as positive examples, while incorrect answers are treated as negative examples. This results in a dataset of 300k positive-negative pairs, which we use to train the scalar RM with Llama3-8B-Instruct.
    \item Synthesis-Based Methods: LMCOR is a synthesis-based method designed for inference scaling. It generates candidate answers by synthesizing responses based on the gold answers in the dataset. We employ Llama3-8B-Instruct as the base model and Llama3.1-70B-Instruct as the policy model to generate candidate responses, keeping all other configurations consistent with the original method.
\end{itemize}

\subsection{Training Implementation}
\label{app:training_details}

For Scalar RM, LMCOR, and our method, we all use Llama-3-8B-Instruct as the base model. 

For our method, we utilize Llama-3-8B-Instruct to generate five candidate responses for each sample. Subsequently, these candidates are subjected to CoT synthesis using Llama3.1-70B-Instruct. For the MATH training dataset, we perform an initial sampling of 50 iterations per sample. If none of these iterations yield a correctly synthesized answer, we apply LLM Repair and conduct a second round of 20 additional samplings. For the WikiTQ training dataset, we carry out an initial sampling of 20 iterations, followed by a second round of 10 additional samplings for samples that fail to produce a correctly synthesized answer in the first round. 
For the data filtering in the TableQA task, we use CritiqueLLM~\cite{ke2023critiquellm}, which is an LLM focused on evaluating the reasoning process and the consistency of answers with the benchmark set through scoring. We retained results with scores greater than or equal to 8 in order to obtain high-quality answers.
Through this data synthesis pipeline, we have generated a comprehensive dataset consisting of 294k synthesized samples for the MATH dataset and 87k synthesized samples for the WikiTQ dataset. Detailed information is shown in Table~\ref{tab:benchmark_sizes}. To ensure the reliability of our results, each experiment is conducted three times, and we report the average outcomes.

Following an extensive hyperparameter search over learning rates, we determine that an LR range of [5e-6, 2e-6, 5e-7] yields optimal performance for all training methods. Thus we choose LR = 2e-6 for all methods to obtain the best outcomes. We utilize the AdamW optimizer~\cite{loshchilov2017decoupled} with decoupled weight decay regularization, setting the weight decay to 1e-2. Dropout is not applied during training.

In our experimental setup, we configure the per-GPU batch size to 1, with gradient accumulation over 16 steps, thereby achieving an effective batch size of 128 when distributed across 8 GPUs, each for 2 epochs. We set the maximum length of 4096 tokens and employ BF16 precision.

\section{Model Definition}
\label{app:glossary}
In this section, we provide detailed definitions and explanations of different LLMs for clarity. Specifically, in the data generation pipeline, the Sampling LLM and Response LLM are used to generate synthetic data, while the Policy Model and Base Model are employed during the Synthesizer's inference stage, as illustrated in Figure~\ref{fig:overview fig}. Below, we formally define these different LLMs:
\begin{itemize}
    \item Policy Model:
    In Synthesizer inference stage, it generates candidate responses to user queries. As shown in Table~\ref{tab:Main results}, the Policy Model can be either API-based (e.g., GPT-4o and GLM-4-Plus) or open-source LLMs (e.g., Llama3-8B-Instruct and Qwen2-7B-Instruct).

    \item Base Model:  
    In Synthesizer inference stage, it is used to train the \model. For simplicity, we use Llama3-8B-Instruct as the Base Model, which is also used as the Sampling LLM in our experiments.

    \item Sampling LLM:  
    In data generation pipeline, it generates diverse CoT candidate responses for queries. We use Llama3-8B-Instruct for this step. These candidates play a crucial role in subsequent processing.

    \item Response LLM:  
    In data generation pipeline, it synthesizes candidate responses into a final answer. We use Llama3.1-70B-Instruct for this step. The final correct answers will serve as training data. Our experiments show that even though the \model is trained with data from only a few models (Llama3-8B-Instruct, Llama3.1-70B-Instruct), it effectively enhances the performance of various Policy Models, indicating strong generalization capabilities.
\end{itemize}

\onecolumn

\section{Prompt for Inference}
\label{app:prompt}
\begin{tcolorbox}[title = {Prompt for \model inference}]

\textbf{[Instruction]}\\
Please act as an excellent summarizer and summarize the following AI responses to the questions. Your summary should fully consider the connection between the question and AI responses, resulting in a correct, high-quality answer. In most cases, the same response that appears most often in the response may be the correct answer. If you find that there is no correct answer, please try to generate a correct answer yourself. Do not copy The candidate's answer, give your summarized answer and reasons, and give the correct answer at the end of the sentence in the format: The answer is...\\

\textbf{[The Start of Original Question]\\}
\{\textcolor{purple}{question}\}\\
\textbf{[The End of Original Question]\\}

\textbf{[The Start of AI Responses]\\}
\{\textcolor{purple}{responses}\}\\
\textbf{[The End of AI Responses]}
\label{box:synthesizer_inference}
\end{tcolorbox}

\begin{tcolorbox}[title = {Prompt for LLM Repair}]
[Instruction]\\
Please act as an excellent summarizer. Most of the candidate answers provided are wrong; you should find the right steps in these AI responses and try to combine these right steps to generate a new correct, high-quality answer. Give your summarized answer and reasons, and give the correct answer at the end of the sentence, in the format: The answer is...

\textbf{[The Start of Original Question]\\}
\{\textcolor{purple}{question}\}\\
\textbf{[The End of Original Question]\\}

\textbf{[The Start of AI Responses]\\}
\{\textcolor{purple}{responses}\}\\
\textbf{[The End of AI Responses]}
\end{tcolorbox}

\begin{tcolorbox}[title = {Prompt for TableQA CoT-prompt}]

Offer a thorough and accurate solution that directly addresses the Question outlined in the [Question]. Make sure that the last sentence is formatted as "the answer is XXX" and try to use words from the table or the question.\\

\#\#\# [Table]: \\
\{\textcolor{purple}{table}\}\\

\#\#\# [Question]: \\
\{\textcolor{purple}{question}\}\\

\#\#\# [Solution]:
\end{tcolorbox}

\begin{tcolorbox}[title = {Prompt for MATH500 CoT-prompt}]

As an expert problem solver, solve step by step the following mathematical questions. Make sure that the last sentence is formatted as "The answer is XXX" and the value of the correct answer is surrounded by boxed{} when giving your answer.\\

[Problem]:\\
\{\textcolor{purple}{question}\}\\

[Solution]:
\end{tcolorbox}

\begin{tcolorbox}[title = {Prompt for GSM8k CoT-prompt}]

As an expert problem solver solve step by step the following mathematical questions. Make sure that the last sentence is formatted as "The answer is XXX" when giving your answer.\\

Q:
\{\textcolor{purple}{question}\}\\
A:
\end{tcolorbox}

\begin{figure*}
    \centering
    \begin{mdframed} [style=exampledefault,frametitle={An example of synthesizing a correct answer from all the wrong candidate responses}]
    \small
\textbf{[Instruction]}

Please act as an excellent summarizer. Below are some candidate answers generated by AI assistants. You need to do two tasks based on these responses: First, you should analyze these AI responses and find the right steps among them. Second, you should try to synthesize these right steps to generate a new correct, high-quality answer. If you can't get a correct answer based on these answers, you should generate a correct answer. Give your summarized answer and reasons, and give the correct answer at the end of the sentence, in the format: The answer is...\\

[The Start of Original Question]\\
As an expert problem solver, solve step by step the following mathematical questions. Make sure that the last sentence is formatted as "The answer is XXX" and the value of the correct answer is surrounded by boxed{} when giving your answer.\\

\#\#\# Problem:\\

What real values of $x$ are not in the domain of\\

$f(x)=\frac{1}{|x^2+3x-4|+|x^2+9x+20|}$?\\

\#\#\# Solution:\\

[The End of Original Question]\\

[The Start of AI Responses]\\

Response 1:...... The answer is \textcolor{red}{$\boxed{-2, -4}$.}\\

Response 2:...... The answer is \textcolor{red}{\boxed{\text{The domain of }f(x)\text{ is the set of all real numbers.}}}\\



Response 3:...... The answer is \textcolor{red}{$\boxed{\{-4,-5\}}$.}\\











Response 4:...... The answer is \textcolor{red}{\boxed{-1, -4, -5, 4}.}\\

Response 5:...... The answer is \textcolor{red}{$\boxed{-4}$.}\\

[The End of AI Responses]

\vspace{1em}

\hrule 

\vspace{1em}

\textbf{Response:\\}

Based on the given AI responses, the correct answer can be determined by analyzing the connection between the question and the responses. The question asks for the real values of x that are not in the domain of the function f(x). \\

Response 1 and Response 5 provide the correct approach to solve the problem by setting the expression inside the absolute value symbols equal to zero and solving for x. However, Response 5 incorrectly states that x = 1 is a solution, which is not supported by the other responses.\\

Response 2 incorrectly states that the domain of f(x) is the set of all real numbers, which is not true since the function has specific values that make the expression inside the absolute value symbols equal to zero.\\

Response 3 and Response 4 provide similar solutions, but Response 4 incorrectly includes x = 4 and x = -1 as solutions, which are not supported by the other responses.\\

Considering the connection between the question and the responses, the correct answer is the values of x that make the expression inside the absolute value symbols equal to zero, which are x = -4 and x = -5 (from Response 3 and Response 4) and x = -2 and x = -4 (from Response 1). However, Response 3 and Response 4 do not agree on the value x = -5, and Response 1 and Response 5 do not agree on the value x = -2 and x = 1, respectively. Since Response 1, Response 3, and Response 5 provide a clear solution, the correct answer is the intersection of the values provided by these responses.\\

The answer is \textcolor{darkgreen}{$\boxed{-4}$}.\\

    \end{mdframed}
      \caption{An example of synthesizing a correct answer from all the wrong candidate responses generated by Llama3-8B-Instruct on MATH500. The ellipsis indicates responses to the MATH instruction by CoT-prompting.}
    \label{fig:case_study}
\end{figure*}

\begin{figure*}
    \centering
    \begin{mdframed} [style=exampledefault,frametitle={An example of synthesizing a correct answer from all the wrong candidate responses}]
    \small
\textbf{[Instruction]}

Please act as an excellent summarizer. Below are some candidate answers generated by AI assistants. You need to do two tasks based on these responses: First, you should analyze these AI responses and find the right steps among them. Second, you should try to synthesize these right steps to generate a new, correct, high-quality answer. If you can't get a correct answer based on these answers, you should generate a correct answer. Give your summarized answer and reasons, and give the correct answer at the end of the sentence in the format: The answer is...\\

[The Start of Original Question]\\
Offer a thorough and accurate solution that directly addresses the Question outlined in the [Question]. Make sure that the last sentence is formatted as "the answer is XXX" and try to use words from the table or the question.\\

\#\#\# [Table Description]\\
The table consists of match records for a certain event, featuring columns for Match ID, Date, Venue, Opponents, and Score. The Match ID includes a code representing the stage of the tournament and a sequential number. The Date column indicates the year of the match, and the Venue column contains the location of the match. The Opponents column lists the competing teams, while the Score column captures the outcome of each match.
\\

\#\#\# [Table]\\
Match,Date,Venue,Opponents,Score\\
GL-A-1,2006..,[[]],[[]],-\\
GL-A-2,2006..,[[]],[[]],-\\
GL-A-3,2006..,[[]],[[]],-\\
GL-A-4,2006..,[[]],[[]],-\\
GL-A-5,2006..,[[]],[[]],-\\
GL-A-6,2006..,[[]],[[]],-\\
Quarterfinals-1,2006..,[[]],[[]],-\\
Quarterfinals-2,2006..,[[]],[[]],-\\
Semifinals-1,2006..,[[]],[[]],-\\
Semifinals-2,2006..,[[]],[[]],-\\

\#\#\# [Question]\\
How many matches are from 2006?\\

[The End of Original Question]\\

[The Start of AI Responses]\\

Response 1: To determine the number of matches from 2006, we need to examine the 'Date' column in the given table. After reviewing the table, we find that all the matches listed have a date in 2006. Specifically, there are 12 matches with dates in 2006, as indicated by the 'GL-A' matches and the tournament stages listed in the 'Match' column. Therefore, the answer is \textcolor{red}{12} matches are from 2006. \\

Response 2: To find the number of matches from 2006, we need to count the rows where the Date column contains the year 2006. In the provided table, there are 10 rows, and 8 of those rows have the Date column indicating the year 2006. Therefore, the answer is \textcolor{red}{8} matches are from 2006.\\

Response 3: ......The answer is \textcolor{red}{8}.\\

Response 4:......Therefore, the answer is \textcolor{red}{8} matches are from 2006.\\

Response 5:......The number of matches from 2006 is \textcolor{red}{8}.\\

[The End of AI Responses]

\vspace{1em}

\hrule 

\vspace{1em}

\textbf{Response:\\}

After reviewing the AI responses, it is clear that the majority of the responses indicate that there are 8 matches from 2006. However, upon closer inspection of the provided table, all rows listed have a date in 2006, which totals 10 matches (including the 8 matches from the "GL-A" series and the 2 quarterfinal matches). The discrepancy in the AI responses might be due to an oversight in counting the quarterfinal matches.\\

The answer is \textcolor{darkgreen}{10} matches are from 2006.

\end{mdframed}
\caption{An example of synthesizing a correct answer from all the wrong candidate responses generated by Qwen2-7B-Instruct on WikiTQ. The ellipsis indicates responses to the WikiTQ instruction by CoT-prompting.}
    \label{fig:case_study_2}
\end{figure*}

\twocolumn

\end{document}